%% file: main.tex
\documentclass[10pt,conference]{IEEEtran}

\usepackage{graphicx}
\usepackage{booktabs}
\usepackage{multirow}
\usepackage{amssymb}
\usepackage{color}
\usepackage{pifont}% http://ctan.org/pkg/pifont
\usepackage{tabularx}
\usepackage{graphics}
\usepackage{makecell}
\usepackage{balance}
\usepackage{hyperref}
\usepackage{textcomp}

\newcommand{\lab}[1]{\textquotesingle{#1}\textquotesingle}

\usepackage{adjustbox}

\usepackage{enumerate}
\usepackage[shortlabels]{enumitem}

\usepackage{xspace}
\newcommand{\tool}[1]{\textsc{#1}\xspace}
\newcommand{\neurospf}{\tool{NeuroSPF}}
\usepackage{listings, xcolor}
\usepackage{pgfplots}
\pgfplotsset{width=7cm,compat=1.8}

\usepackage{tcolorbox}
\usepackage{verbatim}
\usepackage{array}
\newcommand{\PreserveBackslash}[1]{\let\temp=\\#1\let\\=\temp}

\newcolumntype{C}[1]{>{\PreserveBackslash\centering}p{#1}}
\newcolumntype{R}[1]{>{\PreserveBackslash\raggedleft}p{#1}}
\newcolumntype{L}[1]{>{\PreserveBackslash\raggedright}p{#1}}
\makeatletter
\newcommand{\linebreakand}{%
  \end{@IEEEauthorhalign}
  \hfill\mbox{}\par
  \mbox{}\hfill\begin{@IEEEauthorhalign}
}
\makeatother

\title{\neurospf: A tool for the Symbolic\\ Analysis of Neural Networks}

\author{\IEEEauthorblockN{Muhammad Usman}
\IEEEauthorblockA{
\textit{University of Texas at Austin, USA}\\
muhammadusman@utexas.edu}
\and
\IEEEauthorblockN{Yannic Noller}
\IEEEauthorblockA{
\textit{National University of Singapore}\\
yannic.noller@acm.org}
\and
\IEEEauthorblockN{Corina S. P\u{a}s\u{a}reanu}
\IEEEauthorblockA{
\textit{Carnegie Mellon University, KBR Inc., NASA Ames}\\
corina.s.pasareanu@nasa.gov}
\linebreakand
\IEEEauthorblockN{Youcheng Sun}
\IEEEauthorblockA{
\textit{Queen's University Belfast, UK}\\
youcheng.sun@qub.ac.uk}
\and
\IEEEauthorblockN{Divya Gopinath}
\IEEEauthorblockA{
\textit{KBR Inc., NASA Ames}\\
divya.gopinath@nasa.gov}
}

\begin{document}
\maketitle
\input{abstract}

\input{introduction}
\input{background}
\input{tooldescription}
\input{evaluation}
\input{conclusion}

	\bibliographystyle{IEEEtran}
\bibliography{all}

\end{document}

%% file: abstract.tex
\begin{abstract}
This paper presents \neurospf, a tool for the symbolic analysis of neural networks. Given a trained neural network model, the tool extracts the architecture and model parameters and translates them into a Java representation that is amenable for analysis using the Symbolic PathFinder symbolic execution tool. Notably, \neurospf encodes specialized peer classes for parsing the model's parameters, thereby enabling efficient analysis. With \neurospf the user has the flexibility to specify either the inputs or the network internal parameters as symbolic,  promoting the application of program analysis and testing approaches from software engineering to the field of machine learning. For instance, \neurospf can be used for coverage-based testing and test generation, finding adversarial examples and also constraint-based repair of neural networks, thus improving the reliability of neural networks  and of the applications that use them. 
Video URL: \textcolor{blue}{\url{https://youtu.be/seal8fG78LI}}

\end{abstract}

%% file: introduction.tex
\section{Introduction}
\label{sec:introduction}
\input{code/framework}
Deep Neural Networks (DNNs) have gained popularity in recent years, being used in a variety of applications including banking, health-care,  image and speech recognition, and perception in self-driving cars. With this widespread use of DNNs also come serious safety and security concerns. As a result, several techniques for testing \cite{pei2017deepxplore,tian2018deeptest,10.1007/978-3-030-16722-6_10} and verification \cite{huang2017safety,katz2017reluplex,pulina2010abstraction} of neural networks have been developed recently, the majority of which have built dedicated tools.

In this work, we take a different approach, and we present \neurospf, which builds on a mature, widely used, program analysis tool, namely,  Symbolic PathFinder (SPF) \cite{SPF}, to support analysis of neural networks, while leveraging the techniques that are already incorporated in SPF. 

SPF \cite{SPF} combines symbolic execution \cite{King1976} 
with model checking for automated test case generation and error detection in Java byte-code programs. It supports both classical as well as concolic execution, it measures coverage and it is integrated with different constraint solvers, implementing also incremental analysis and solving -- all these features could be useful for the analysis of neural networks as well. \neurospf extends SPF to support analysis of neural network models {\em efficiently}. To this end, \neurospf first translates a trained neural network model specified in Keras into Java and uses specialized  peer  classes  to  enable  efficient  parsing  of the  model’s  parameters. 
Furthermore, \neurospf enables users to make both the network inputs (e.g., input pixels for an image classifier)  and the network parameters (i.e., weights and biases) symbolic, via special annotations. This flexibility opens up the possibility for several interesting applications.

For instance, \neurospf can be used for testing and test input generation with respect to coverage criteria that are relevant for neural networks \cite{pei2017deepxplore,ma2018deepgauge,sun2019structural,kim2019guiding,feng2020deepgini}. This can be achieved by solving the relevant constraints collected by \neurospf. \neurospf can also be used for analyzing the robustness and for generating adversarial examples in neural networks, as studied in \cite{Gopinath2019SymbolicEF,gopinath2019symbolic,sun2018concolic}, which all propose specialized symbolic execution techniques for adversarial testing. Furthermore, the symbolic analysis in \neurospf can enable the automatic inference of neural network properties as advocated by Gopinath et al. \cite{8952519}. These properties are network preconditions built based on the constraints collected with a symbolic analysis of the network. We also envision that \neurospf can enable automated repair for neural networks, by leveraging existing constraint-based repair techniques from the software engineering community \cite{10.5555/2486788.2486890,ma2018mode,islam2020repairing,sohn2019search} and adapting them to the specifics of neural networks. We summarize our contributions as follows:

\begin{itemize}
    \item We present \neurospf, a tool which facilitates the symbolic analysis of neural networks; \neurospf can handle feed-forward neural networks with dense, convolutional, and pooling layers, with ReLU activations and Softmax functions. 
    \item To achieve efficient analysis, \neurospf encodes specialized peer classes for parsing and storing the model's parameters. 
    \item We evaluate \neurospf on three neural networks (MNIST low quality, MNIST high quality, and CIFAR-10), showcasing \neurospf's ability of handling complex neural network models and highlighting the importance of using the peer classes. 
    \item We also provide a detailed demonstration on how to use \neurospf, illustrating robustness analysis for a neural network model trained on the MNIST dataset. 
\end{itemize}

The {\bf envisioned users} for \neurospf include researchers and software engineers interested in applying symbolic execution for testing and debugging neural network models.
The {\bf challenge} we propose to address stems from the need to better understand and debug neural networks which are essentially black boxes. The {\bf methodology} it implies for its users is described in detail in Section \ref{sec:tool}. The results of preliminary {\bf validation} are described in Section \ref{sec:evaluation}. We further plan for in-depth studies on the use of \neurospf in the applications that we outlined: testing, attack generation, property inference and automated repair for neural networks.

%% file: code/framework.tex
\begin{figure*}[t]
    \centering
    \includegraphics[scale=0.4]{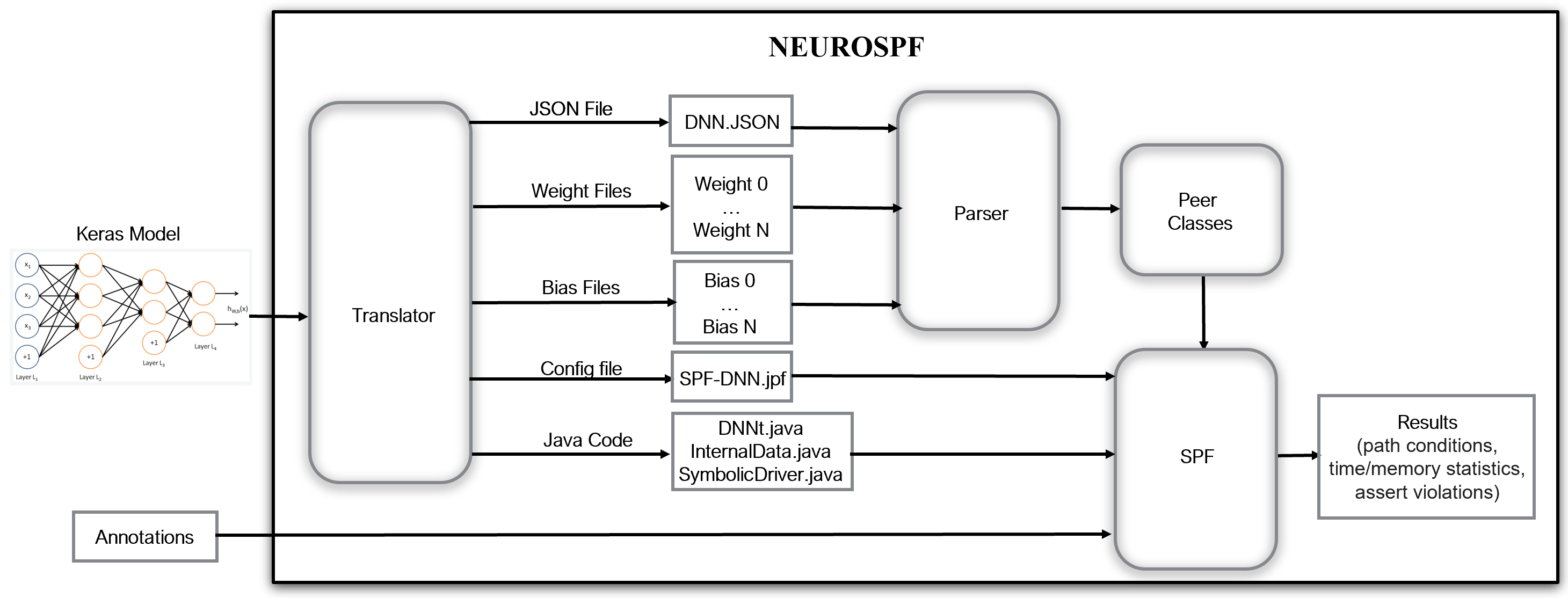}
    \caption{Overview of \neurospf}
    \label{fig:overview}
    \vspace{-4mm}
\end{figure*}

%% file: background.tex
\section{Background}
\label{sec:background}

\subsection{Neural Networks}

Neural networks (NNs) \cite{Goodfellow-et-al-2016} are machine learning algorithms that can be trained to perform different tasks such as classification and regression.
NNs consist of multiple layers, starting from the \textit{input} layer, followed by one or more \textit{hidden} layers (such as convolutional, dense, activation, and pooling), and a final \textit{decision} layer.
Each layer consists of a number of computational units, called \textit{neurons}. Each neuron applies an activation function on a weighted sum of its inputs;
$N(X)=\sigma(\sum_i w_i \cdot N_i(X) + b)$
where $N_i$ denotes the value of the $i^{th}$ neuron in the previous layer of the network and the coefficients $w_i$ and the constant $b$ are referred to as \emph{weights} and \emph{bias}, respectively; $\sigma$ represents the activation function.
For instance, the ReLU (rectified linear unit) activation function returns its input as is if it is positive, and returns 0 otherwise, i.e., $\sigma(X) = max(0, X)$.
The final decision layer (\textit{logits}) typically uses a specialized function (e.g., max or \textit{softmax}) to determine the decision or the output of the network.

\subsection{Symbolic Execution and Symbolic PathFinder }
In symbolic execution \cite{King1976} a program is executed with symbolic (i.e., unspecified) inputs rather than concrete inputs.
The goal is to generate mathematical constraints from the conditions in the program, which can be solved to generate test inputs.
Symbolic PathFinder (SPF) \cite{SPF} builds on top of Java PathFinder model checker to enable symbolic execution of Java bytecode programs.
SPF can perform both standard symbolic execution and concolic execution, by collecting symbolic constraints along concrete executions. Both can be leveraged for neural network analysis \cite{Gopinath2019SymbolicEF,gopinath2019symbolic,sun2018concolic}.

%% file: tooldescription.tex
\section{Tool Description}
\label{sec:tool}

\subsubsection{Methodology}

Figure~\ref{fig:overview} shows the overall framework of \neurospf. Users need to input the Keras model into the \neurospf. The \emph{translator} component will generate:  (1) a \emph{}{JSON file} that provides critical information about the network model such as the dimensions of the weights and biases, number and types of layers, (2) \emph{Weights and Biases files} that contain the values of the weights and biases for all layers, (3) \emph{Config file} (SPF-DNN.jpf) which contains configuration settings for SPF i.e., minimum and maximum range of symbolic variables, type of constraint solver to be used etc. and (4) the \emph{Java code} representation of Keras model.

The JSON file along with weights and biases files are the inputs to the \emph{Parser} component. The parser reads the dimensions of each layer from the JSON file and loads the weights and biases from the respective files via specialized \emph{Peer} classes. Once the files have been read via \emph{Peer} classes, SPF is executed. The Java code along with the Config file form the input to SPF. The user has the option to edit the Config file according to their requirements. 

The analysis can be further configured via user {\em annotations} which specify which inputs or parameters to be considered symbolic. Such a selection can be done based on attribution methods as described in previous work~\cite{Gopinath2019SymbolicEF}.
The output of the tool consists of the analysis results computed by SPF (assert violations, coverage information, time and memory statistics).
\input{code/activationlayer}
\input{code/jpfconfig}
\input{code/symbolicdriver}
\input{code/adversarialinput}
\input{code/adversarialimage}

\subsubsection{Translation to Java} Figure~\ref{fig:activationlayer} shows the Java representation of an activation layer with ReLUs. Layer 7 is an array with size \emph{128}. The \emph{for loop} traverses through the output of previous layer i.e., layer 6. If the output is greater than 0, the neuron is activated by setting it to the output value of the corresponding neuron of previous layer otherwise the value is set to 0. 

The \emph{translator} creates an \emph{InternalData} class 
that contains arrays to store the weights and biases of neural network layers. \emph{InternalData} also provides a function to read weights/biases using I/O libraries. This is helpful if the user wants to use the standalone Java code to run the neural network model without symbolic execution. Otherwise, \emph{SymbolicDriver} reads the weights/biases files using specialized \emph{Peer} classes for efficient symbolic execution. A JSON file is also generated. It encodes information about the architecture of the model.

As mentioned, the \emph{translator} also generates a Config file (SPF-DNN.jpf) to specify configuration settings for SPF. Figure~\ref{fig:jpfconfig} shows a sample Config file. Line \emph{1} specifies the target class to be executed using SPF. Lines \emph{2} and \emph{3} specify the classpath and sourcepath respectively. Lines \emph{4} and \emph{5} specify the minimum and maximum range of symbolic variables i.e., 0 and 255 respectively. Line \emph{6} specifies the type of constraint solver to be used, i.e.,  Z3 \cite{MouraBjorner08Z3}.

\subsubsection{Parser and Peer Classes}
There are three inputs to the \emph{Parser} component, i.e., the JSON file (which contains the description of the architecture of the neural network), weights and biases files for the neural network. The \emph{parser} reads the JSON file and loads data from weights and biases files. \neurospf reads these files via \emph{Peer} classes in SPF. This is a mechanism for executing Java code in the native VM instead of SPF's custom VM, which is much more efficient  than to read files directly, as I/O operations significantly slow down the symbolic execution in SPF (see the next section for a comparison between \neurospf and plain (Vanilla) SPF). \emph{Peers} avoid the bytecode interpretation in the custom VM and directly execute the Java bytecode.

\emph{DNNLayer} is an abstract class and there are concrete classes for each type of layers. For example, \emph{ActivationLayer}, \emph{ConvolutionLayer} and \emph{DenseLayer} are all included in \neurospf. There are specific member variables and methods for these layers depending on their functionality. This information is filled by parsing the JSON file.
Specific layers for other neural network types can be added as needed. \neurospf currently supports Keras models but we plan to add support for other machine learning libraries in the future.

\subsubsection{Symbolic Driver} The symbolic driver provides the main entry point for running the neural network. Figure~\ref{fig:symbolicdriver} shows an example \emph{SymbolicDriver} created for an image classifier. 

%% file: code/activationlayer.tex
\begin{figure}[t]
    \centering
  \begin{lstlisting}[language = Java, xleftmargin=5.0ex, frame = single ,firstnumber = 1 , escapeinside={(*@}{@*)}]
double[] layer7=new double[128];
for(int i=0; i<128; i++)
    if(layer6[i]>0) layer7[i]=layer6[i];
    else layer7[i]=0;
\end{lstlisting}
    \caption{DNNt.java - Activation Layer (ReLU)}
    \label{fig:activationlayer}
    \vspace{-5mm}
\end{figure}

%% file: code/jpfconfig.tex
\begin{figure}[t]
    \centering
  \begin{lstlisting}[language = Java, frame = single ,firstnumber = 1 ,xleftmargin=5.0ex, escapeinside={(*@}{@*)}]
target=neurospf.SymbolicDriver
classpath=${jpf-symbc}/build/examples/
sourcepath=${jpf-symbc}/src/examples/
symbolic.min_double=0.0
symbolic.max_double=255.0
symbolic.dp=z3
\end{lstlisting}
\vspace{-3mm}
    \caption{Config File (SPF-DNN.jpf)}
    \label{fig:jpfconfig}

\end{figure}

%% file: code/symbolicdriver.tex
\begin{figure}[t]
    \centering
  \begin{lstlisting}[language = Java, frame = single ,firstnumber = 1 ,xleftmargin=3ex, escapeinside={(*@}{@*)}]
Method to load image in image[28][28] array
InternalData internaldata = new InternalData();    
DNNGeneralize.readDataFromFiles(path+"params\\",path+"dnn.json");
internaldata.biases0 = (double[]) DNNGeneralize.get_data("biases0");
...
internaldata.weights0 = (double[][][][]) DNNGeneralize.get_data("weights0");
...
DNNt model=new DNNt(internaldata);
int label = model.run(image);
\end{lstlisting}
\vspace{-3mm}
    \caption{SymbolicDriver.java}
    \label{fig:symbolicdriver}
    \vspace{-5mm}
\end{figure}

%% file: code/adversarialinput.tex
\begin{figure}[t]
    \centering
  \begin{lstlisting}[language = Java, xleftmargin=5.0ex, frame = single ,firstnumber = 1 , escapeinside={(*@}{@*)}]
image[15][15][0]=Debug.addSymbolicDouble(image[15][15][0],"sym_15_15");
.... 
if(label!=8) {
   Debug.getSolvedPC();
   Debug.getSymbolicRealValue(image[15][15][0]);
    assert(false);}
\end{lstlisting}
    \caption{SymbolicDriver.java - Code for Adversarial Generation}
    \label{fig:adversarialinput}
\end{figure}

%% file: code/adversarialimage.tex
\begin{figure}[]
    \centering
    \begin{tabular}{@{}c@{\hspace{0.2cm}}c@{\hspace{1.0cm}}c@{\hspace{0.2cm}}c@{}}   
    \includegraphics[]{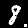} &  \includegraphics[]{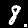}\\
     {\lab 8}    &       {\lab 9}
    \end{tabular}
    \caption{ 
    Label {\lab 8} $\rightarrow$  {\lab 9}, by changing pixel([15][15]) value from  0  to  218}
    \label{fig:adversarial-examples}
    \vspace{-4mm}
\end{figure}

%% file: evaluation.tex
\section{Evaluation}
\label{sec:evaluation}

In this section, we present the  application of \neurospf in the popular field of adversary generation for neural network models. Specifically, we demonstrate how \neurospf can be used to generate adversarial examples for a neural network model trained on MNIST dataset (henceforth named MNIST-LowQuality). 
We also measure the run-time overhead incurred by the Java translation on neural networks trained for image classification (using MNIST and CIFAR-10 data sets) and compare the performance (execution time) of Vanilla SPF (SPF without the peer classes) and \neurospf.

Experiments  were  performed  on  a  Windows  10.0 with Intel Core-i9 and 64GB RAM. The \neurospf tool along with neural network models are available at \textcolor{blue}{\url{https://github.com/muhammadusman93/neurospf}}.

\subsection{Robustness Analysis}

We analyzed a trained MNIST model with \neurospf using a randomly selected input image (784 pixels). The image is represented using a 2D array of size [28][28]. For illustration purposes, we made one pixel (15,15) symbolic and added an assertion that triggers an error when the output class is not {\lab 8}.  Figure~\ref{fig:adversarialinput} shows the sample annotations for making one pixel symbolic. \neurospf took 54 seconds to generate a counterexample, i.e., an adversarial image that led the neural network to change its output to label {\lab 9} from label {\lab 8}. Figure \ref{fig:adversarial-examples} shows the adversarial input found.
This simple example demonstrates how \neurospf can be used to assess the neural network model's robustness to adversarial perturbations and to generate adversarial examples. A similar analysis is described in \cite{gopinath2019symbolic}, which uses a dedicated symbolic execution tool, and modifies a small set of pixels (in some cases one or two). These pixels are discovered with an attribution analysis which can potentially be leveraged in \neurospf as well. 

\subsection{Measuring Run-time Overhead}

Table~\ref{tab:comparison} compares the performance (execution time) of the following: (1) Keras model representation, (2) Java code representation, (3) Vanilla SPF (SPF without peer classes) and (4) \neurospf. Our experiments are based
on the commonly used datasets MNIST and CIFAR-10. The MNIST models are 10-layer convolutional neural networks (CNNs) and have the typical structure of modern neural networks such as convolutional, dense, max-pooling and softmax layers. The CIFAR-10 model is a 15-layer CNN with 890k trainable parameters. The results show that all 4 settings give the same accuracy confirming the correctness of our implementation. 

For MNIST-LowQuality model, the Keras representation is able to predict labels for 100 images in 0.1 seconds whereas the Java representation takes 0.2 seconds. Vanilla SPF times out while \neurospf takes just 34.1 seconds. 
For MNIST-HighQuality model, the Keras representation is able to predict labels for 100 images in 0.1 seconds whereas Java representation takes 0.2 seconds. Vanilla SPF takes 2424.3 seconds (40 minutes)  while \neurospf takes just 43.3 seconds. 
For CIFAR-10 model, the Keras representation is able to predict labels for 100 images in 0.1 seconds whereas Java representation takes 4.2 seconds. Vanilla SPF times out while \neurospf takes 1908.1 seconds (32 minutes). 

As expected, the Keras representation outperforms the Java representation in execution time but the major benefit of the Java representation is that existing software testing and verification techniques and tools for Java can be applied to deep learning models, with reasonable effort. Our results also show that \neurospf significantly outperforms the original SPF in terms of execution time. This is because \neurospf encodes specialized peer classes for efficient parsing of NN model parameters. 
\input{table1}

The results indicate that \neurospf is effective in executing neural networks with complex features (convolutional, dense, max-pooling, ReLU and softmax layers). They do not give an indication of the scalability of a symbolic analysis, which depends on the number of variables that are marked as symbolic and the number of symbolic paths and constraints that are generated.

We view \neurospf as an open source platform for researchers and software engineers who want to experiment with different symbolic analysis on neural networks using familiar languages and techniques. Simply running symbolic execution over the whole network (with all inputs symbolic) will likely not scale and specialized heuristics will be needed to enable \neurospf to perform specific analyses.

%% file: table1.tex
\begin{table}
	\centering
	\caption{Comparison between Keras Model representation, Java code representation, Vanilla SPF and \neurospf; ``-" indicates time-out of 1 Hour}
\label{tab:comparison}
		\begin{tabular}{|L{1.8cm}|R{0.6cm}|C{0.6cm}|C{0.6cm}|C{0.7cm}|R{1.3cm}|}
		 \hline
		  
		   	      \multirow{2}{*}{\textbf{Model}} & \multirow{1}{*}{\textbf{Acc}} &
\multicolumn{4}{c|}{\textbf{Time (s)}} \\ \cline{3-6}	
    
		   	    \textbf{}  &  \textbf{\%} & \textbf{Keras}  & \textbf{Java} & \textbf{SPF}  & \textbf{\neurospf} \\	\hline
            MNIST-LowQ       &96.0  &0.1 & 0.2    &-      &34.1  \\
            MNIST-HighQ       &100.0 &0.1 & 0.2    &2424.3 &43.3  \\
            CIFAR-10     &87.0  &0.1 & 4.2    &-      &1908.1  \\
          \hline

               \hline

\end{tabular}
\vspace{-3mm}
\end{table}

%% file: conclusion.tex
\section{Conclusion}
\label{sec:conclusion}

We presented the \neurospf tool that analyzes neural networks using Symbolic PathFinder. We showed how \neurospf can be used to check for adversarial robustness in neural networks. In the future, we plan to investigate other applications of \neurospf, such as test input generation, property inference and automated constraint-based repair. 

%% file: main.bbl
% Generated by IEEEtran.bst, version: 1.14 (2015/08/26)
\begin{thebibliography}{10}
\providecommand{\url}[1]{#1}
\csname url@samestyle\endcsname
\providecommand{\newblock}{\relax}
\providecommand{\bibinfo}[2]{#2}
\providecommand{\BIBentrySTDinterwordspacing}{\spaceskip=0pt\relax}
\providecommand{\BIBentryALTinterwordstretchfactor}{4}
\providecommand{\BIBentryALTinterwordspacing}{\spaceskip=\fontdimen2\font plus
\BIBentryALTinterwordstretchfactor\fontdimen3\font minus
  \fontdimen4\font\relax}
\providecommand{\BIBforeignlanguage}[2]{{%
\expandafter\ifx\csname l@#1\endcsname\relax
\typeout{** WARNING: IEEEtran.bst: No hyphenation pattern has been}%
\typeout{** loaded for the language `#1'. Using the pattern for}%
\typeout{** the default language instead.}%
\else
\language=\csname l@#1\endcsname
\fi
#2}}
\providecommand{\BIBdecl}{\relax}
\BIBdecl

\bibitem{pei2017deepxplore}
K.~Pei, Y.~Cao, J.~Yang, and S.~Jana, ``{DeepXplore}: Automated whitebox
  testing of deep learning systems,'' in \emph{proceedings of the 26th
  Symposium on Operating Systems Principles}, 2017, pp. 1--18.

\bibitem{tian2018deeptest}
Y.~Tian, K.~Pei, S.~Jana, and B.~Ray, ``{DeepTest}: Automated testing of
  deep-neural-network-driven autonomous cars,'' in \emph{ICSE}, 2018, pp.
  303--314.

\bibitem{10.1007/978-3-030-16722-6_10}
H.~F. Eniser, S.~Gerasimou, and A.~Sen, ``Deep{F}ault: Fault localization for
  deep neural networks,'' in \emph{Fundamental Approaches to Software
  Engineering}, R.~H{\"a}hnle and W.~van~der Aalst, Eds.\hskip 1em plus 0.5em
  minus 0.4em\relax Cham: Springer International Publishing, 2019, pp.
  171--191.

\bibitem{huang2017safety}
X.~Huang, M.~Kwiatkowska, S.~Wang, and M.~Wu, ``Safety verification of deep
  neural networks,'' in \emph{CAV}.\hskip 1em plus 0.5em minus 0.4em\relax
  Springer, 2017, pp. 3--29.

\bibitem{katz2017reluplex}
G.~Katz, C.~Barrett, D.~L. Dill, K.~Julian, and M.~J. Kochenderfer, ``Reluplex:
  An efficient smt solver for verifying deep neural networks,'' in
  \emph{CAV}.\hskip 1em plus 0.5em minus 0.4em\relax Springer, 2017, pp.
  97--117.

\bibitem{pulina2010abstraction}
L.~Pulina and A.~Tacchella, ``An abstraction-refinement approach to
  verification of artificial neural networks,'' in \emph{CAV}.\hskip 1em plus
  0.5em minus 0.4em\relax Springer, 2010, pp. 243--257.

\bibitem{SPF}
C.~S. P\u{a}s\u{a}reanu, W.~Visser, D.~H. Bushnell, J.~Geldenhuys, P.~C.
  Mehlitz, and N.~Rungta, ``{Symbolic PathFinder}: integrating symbolic
  execution with model checking for {Java} bytecode analysis,'' \emph{Autom.
  Softw. Eng.}, vol.~20, no.~3, pp. 391--425, 2013.

\bibitem{King1976}
J.~C. King, ``{Symbolic Execution and Program Testing},'' \emph{Commun. ACM},
  vol.~19, no.~7, pp. 385--394, jul 1976.

\bibitem{ma2018deepgauge}
L.~Ma, F.~Juefei-Xu, F.~Zhang, J.~Sun, M.~Xue, B.~Li, C.~Chen, T.~Su, L.~Li,
  and Y.~Liu, ``{DeepGauge}: Multi-granularity testing criteria for deep
  learning systems,'' in \emph{ASE}, 2018.

\bibitem{sun2019structural}
Y.~Sun, X.~Huang, D.~Kroening, J.~Sharp, M.~Hill, and R.~Ashmore, ``Structural
  test coverage criteria for deep neural networks,'' \emph{ACM TECS}, vol.~18,
  no.~5s, pp. 1--23, 2019.

\bibitem{kim2019guiding}
J.~Kim, R.~Feldt, and S.~Yoo, ``Guiding deep learning system testing using
  surprise adequacy,'' in \emph{ICSE}.\hskip 1em plus 0.5em minus 0.4em\relax
  IEEE, 2019, pp. 1039--1049.

\bibitem{feng2020deepgini}
Y.~Feng, Q.~Shi, X.~Gao, J.~Wan, C.~Fang, and Z.~Chen, ``Deep{G}ini:
  {P}rioritizing massive tests to enhance the robustness of deep neural
  networks,'' in \emph{Proceedings of the 29th ACM SIGSOFT International
  Symposium on Software Testing and Analysis}, 2020, pp. 177--188.

\bibitem{Gopinath2019SymbolicEF}
D.~Gopinath, C.~Pasareanu, K.~Wang, M.~Zhang, and S.~Khurshid, ``Symbolic
  execution for attribution and attack synthesis in neural networks,''
  \emph{ICSE (ICSE-Companion)}, pp. 282--283, 2019.

\bibitem{gopinath2019symbolic}
D.~Gopinath, M.~Zhang, K.~Wang, I.~B. Kadron, C.~Pasareanu, and S.~Khurshid,
  ``Symbolic execution for importance analysis and adversarial generation in
  neural networks,'' in \emph{ISSRE}, 2019, pp. 313--322.

\bibitem{sun2018concolic}
Y.~Sun, M.~Wu, W.~Ruan, X.~Huang, M.~Kwiatkowska, and D.~Kroening, ``Concolic
  testing for deep neural networks,'' in \emph{ASE}, ser. ASE 2018.\hskip 1em
  plus 0.5em minus 0.4em\relax New York, NY, USA: ACM, 2018, p. 109–119.

\bibitem{8952519}
D.~{Gopinath}, H.~{Converse}, C.~{Pasareanu}, and A.~{Taly}, ``Property
  inference for deep neural networks,'' in \emph{ASE}, 2019, pp. 797--809.

\bibitem{10.5555/2486788.2486890}
H.~D.~T. Nguyen, D.~Qi, A.~Roychoudhury, and S.~Chandra, ``{SemFix}: Program
  repair via semantic analysis,'' in \emph{ICSE}, ser. ICSE '13.\hskip 1em plus
  0.5em minus 0.4em\relax IEEE Press, 2013, p. 772–781.

\bibitem{ma2018mode}
S.~Ma, Y.~Liu, W.-C. Lee, X.~Zhang, and A.~Grama, ``{MODE}: automated neural
  network model debugging via state differential analysis and input
  selection,'' in \emph{ESEC/FSE}, 2018, pp. 175--186.

\bibitem{islam2020repairing}
M.~J. Islam, R.~Pan, G.~Nguyen, and H.~Rajan, ``Repairing deep neural networks:
  Fix patterns and challenges,'' \emph{arXiv preprint arXiv:2005.00972}, 2020.

\bibitem{sohn2019search}
J.~Sohn, S.~Kang, and S.~Yoo, ``Search based repair of deep neural networks,''
  \emph{arXiv preprint arXiv:1912.12463}, 2019.

\bibitem{Goodfellow-et-al-2016}
I.~Goodfellow, Y.~Bengio, and A.~Courville, \emph{Deep Learning}.\hskip 1em
  plus 0.5em minus 0.4em\relax MIT Press, 2016,
  \url{http://www.deeplearningbook.org}.

\bibitem{MouraBjorner08Z3}
L.~de~Moura and N.~Bjorner, ``Z3: An efficient {SMT} solver,'' in \emph{TACAS},
  2008.

\end{thebibliography}
